\documentclass[conference]{IEEEtran}

\usepackage{graphicx}
\usepackage{cite}
\usepackage{lettrine}
\def\BibTeX{{\rm B\kern-.05em{\sc i\kern-.025em b}\kern-.08em
    T\kern-.1667em\lower.7ex\hbox{E}\kern-.125emX}}

\usepackage[T1]{fontenc}
\usepackage[hidelinks]{hyperref}
\makeatletter
\newcommand*{\emails}[2][.bee21seecs@seecs.edu.pk]{%
    \def\@tempa{\@gobble}%
    \@for\qrr@email:=#2\do{%
        \edef\@tempb{\noexpand\href{mailto:\qrr@email #1}{\qrr@email}}%
        \edef\@tempa{\unexpanded\expandafter{\@tempa}{, }\unexpanded\expandafter{\@tempb}}}%
    \{\@tempa\}#1%
}

\usepackage{array, booktabs, makecell}
\usepackage[skip=1ex, 
            font=small,labelfont=bf]{caption}
\usepackage{tabularray}
\UseTblrLibrary{amsmath, siunitx}
\usepackage{float}
\usepackage{tikz,xcolor,hyperref}

\definecolor{lime}{HTML}{A6CE39}
\DeclareRobustCommand{\orcidicon}{%
	\begin{tikzpicture}
	\draw[lime, fill=lime] (0,0) 
	circle [radius=0.16] 
	node[white] {{\fontfamily{qag}\selectfont \tiny ID}};
	\draw[white, fill=white] (-0.0625,0.095) 
	circle [radius=0.007];
	\end{tikzpicture}
	\hspace{-2mm}
}

\foreach \x in {A, ..., Z}{%
	\expandafter\xdef\csname orcid\x\endcsname{\noexpand\href{https://orcid.org/\csname orcidauthor\x\endcsname}{\noexpand\orcidicon}}
}

\title{Fault Diagnosis on Induction Motor using Machine Learning and Signal Processing}

\author{Muhammad Samiullah\orcidA{},  Hasan Ali,  Shehryar Zahoor and Anas Ali\\ \textit{School of Electrical Engineering and Computer Science, (SEECS)}\\ \textit{National University of Sciences and Technology} \\ Islamabad, Pakistan \\ \emails{mullah,hali,szahoor,aali}}
\date{}

\begin{document}
\maketitle
\begin{abstract}
    
The detection and identification of induction motor faults using machine learning and signal processing is a valuable approach to avoiding plant disturbances and shutdowns in the context of Industry 4.0. In this work, we present a study on the detection and identification of induction motor faults using machine learning and signal processing with MATLAB Simulink. We developed a model of a three-phase induction motor in MATLAB Simulink to generate healthy and faulty motor data. The data collected included stator currents, rotor currents, input power, slip, rotor speed, and efficiency. We generated four faults in the induction motor: open circuit fault, short circuit fault, overload, and broken rotor bars. We collected a total of 150,000 data points with a 60-40\% ratio of healthy to faulty motor data. We applied Fast Fourier Transform (FFT) to detect and identify healthy and unhealthy conditions and added a distinctive feature in our data. The generated dataset was trained different machine learning models. On comparing the accuracy of the models on the test set, we concluded that the Decision Tree algorithm performed the best with an accuracy of about 92\%. Our study contributes to the literature by providing a valuable approach to fault detection and classification with machine learning models for industrial applications.\\

\end{abstract}


\section{\textbf{Introduction}}
    \lettrine[findent=1pt]{\textbf{T}}{ }he advent of the fourth industrial revolution has made machine-to-machine interaction possible with sophisticated sensors that allow real-time adaption, analysis, and optimum decision-making easy. In the industrial revolution, induction motors as the workhorse of industry are utilized in most industrial processes and domestic applications. They can be found in numerous applications such as blowers, exhaust fans, pumps, and overhead cranes. Due to a harsh operating environment, they are exposed to different conditions, which create faults in them. Continuous monitoring and interaction of motor operating parameters such as vibration, current, and temperature with sensors enable us to diagnose and identify the related issues within interconnected processes and even plan predictive maintenance.
    
    Faults can be avoided and repaired if detected early. In industrial, commercial, and domestic applications, the widely used motors are induction motors due to having less cost and ruggedness. However, a motor can fail suddenly due to non-observance and less maintenance in certain operating conditions. It is reported that 35 to 40 percent of induction motors fail due to electrical faults. These faults include open circuit, short circuit, rotor faults (broken rotor bar) and overloading.

    
    So, from the above, we can see that, induction motors play a crucial role in powering industrial machinery and reliable motor operation is vital for productivity and safety. The motor faults can lead to downtime, increased maintenance costs, and compromised efficiency. The main objective of our project is to develop a system for early fault detection in induction motors and for that we will utilize machine learning and signal processing for accurate and automated fault detection.

\section{\textbf{Literature Review}}
    The literature reveals a growing emphasis on leveraging machine learning and signal processing for fault detection in induction motors \cite{firstline}. Traditionally, condition monitoring in motors relied on methods such as vibration analysis and temperature monitoring. The advent of the fourth industrial revolution has spurred interest in integrating deep learning models for more accurate and automated fault detection. Signal processing techniques, including Fast Fourier Transform, have been pivotal in analyzing motor signals to identify faults \cite{signalprocessing}.  Studies underscore the significance of comprehensive data collection, considering parameters such as stator currents, rotor currents, and efficiency. Comparative analyses of machine learning models highlight the importance of selecting the most suitable approach for fault detection \cite{riera2015advances}. Challenges such as real-time monitoring and model interpretability are acknowledged, emphasizing the need for practical solutions. The severe impact of motor faults on productivity and safety underscores the urgency for early fault detection systems \cite{drif2014stator}. In this context, our project contributes by utilizing machine learning and signal processing in MATLAB Simulink for early and accurate fault detection in induction motors \cite{zhang2016intelligent}.

\section{\textbf{Motor Design and Specifications}}

We have modelled a three phase induction motor in MATLAB Simulink as shown in \textit{figure 1}. 

\begin{figure*}[htp]
    \centering
    \includegraphics[width=\textwidth]{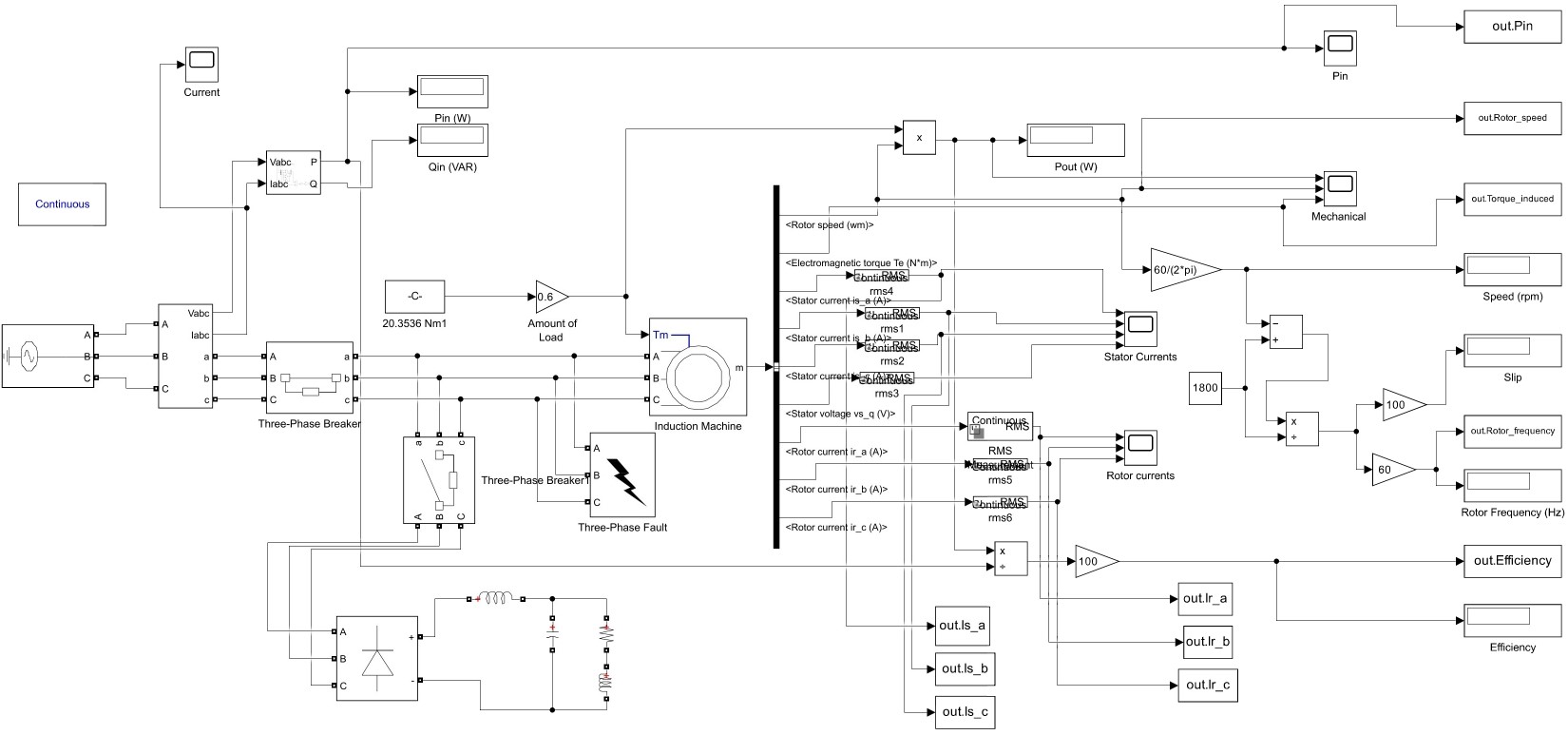}
    \caption{Simulink Model of Three-Phase Induction Motor}
    \label{fig:SM32}
\end{figure*}

In the Simulink model, we are measuring stator and rotor currents, input power, slip rotor speed and efficiency of the three-phase induction motor.  We have used scopes to measure currents, rotor speed and torque and input power which makes it easier for us to understand and monitor the different values. Detailed motor parameters are shown in Table I.

\begin{table}
\centering
\caption{Parameter of Induction Motor}
\label{tab:my_table}
\begin{tabular}{ l  l }
\toprule
Motor Parameters & Value \\
\midrule
Power rating & 5 hp \\

Voltage & 220 V \\

Current & 5.99 A \\

Torque rating & 22.2 N.m \\

Slip & 5.06 \% \\

Frequency & 60 Hz \\

Rated Speed & 1750 RPM \\

\bottomrule
\end{tabular}

\end{table}

Let us see how several faults were induced in the induction motor.

\section{\textbf{Fault Induction in the Motor}}

The open circuit fault is generated by a three phase breaker in the stator lines. It is generated when any of the three phase are open circuited and there is no current flow in the stator winding. 

The short circuit fault is generated by three-phase fault block. This fault causes high amount of currents to flow through the windings representing an inter-turn short-circuiting. The motor cannot withstand such high amount of current and this leads to the breakdown of the motor. 

For every machine, there is a specific rating of it’s application beyond which it cannot operate properly. Similarly, induction motor has its rated torque value. The overload fault is generated when induction motor is loaded above this rating. 

Whenever there is a breakage or crack in the rotor bars of the motor, several harmonics are added to its signals \cite{faiz2014emd}. Hence to model the broken rotor bar fault, we have added harmonics in our model by connecting an RLC circuit to its input line. This would introduce harmonics the pure sinusoidal input signal, hence representing a broken rotor bar fault.

\section{\textbf{Collection of Data}}

In our project we aimed at collecting a total of 150,000 datapoints, with a ratio of 60-40\% of healthy to faulty motors. In simpler terms we gathered 90,000 datapoints for a machine working under normal conditions. We collected majority of data under normal conditions, therefore modelling real world scenario where most times the machine is working under normal condition. Our simulation run time was 5 seconds, each run time generated 10,000 datapoints. Which implies that we ran our simulation of a healthy motor 9 times. The remaining 60,000 data points were collected for faulty machines consisting of 15,000 datapoints for each specific fault. In order to develop an efficient model we collected vast amount of data at different load conditions. We increased the diversity by inducing faults in different phases and varied severity.

\begin{table}
\small
\caption{Generated Dataset}

\sisetup{per-mode = symbol}

\begin{tabular}{ l  l  l  l  l  l  l  l }
\toprule
\thead{Stator\\Curr.} & \thead{Rotor\\Curr. } & \thead{Input\\Power  } & \thead{Rotor\\Freq. } & \thead{Rotor\\Speed } & \thead{Torque\\Ind. } & Effic. & \thead{Fault\\Label}  \\
\midrule
5.53 & 1.31 & 8,489 & 1.11 & 185.01 & 17.35 & 35.49 & 4 \\

5.53 & 1.30 & 8,956 & 1.11 & 185.01 & 17.35 & 33.64 & 4 \\

5.53 & 1.30 & 8,448 & 1.11 & 185.01 & 17.35 & 35.66 & 4 \\

5.53 & 1.29 & 7,417 & 1.11 & 185.01 & 17.35 & 40.61 & 4 \\

5.53 & 1.29 & 6,683 & 1.11 & 185.01 & 17.35 & 45.08 & 4 \\

5.53 & 1.28 & 6,763 & 1.11 & 185.01 & 17.35 & 44.54 & 4 \\

5.53 & 1.28 & 6,781 & 1.11 & 185.01 & 17.35 & 44.43 & 4 \\

5.53 & 1.28 & 6,781 & 1.11 & 185.01 & 17.35 & 44.43 & 4 \\

5.53 & 1.28 & 7,133 & 1.11 & 185.01 & 17.35 & 42.23 & 4 \\

5.53 & 1.28 & 7,830 & 1.11 & 185.01 & 17.35 & 38.47 & 4 \\

5.53 & 1.28 & 8,246 & 1.11 & 185.01 & 17.35 & 36.53 & 4 \\

5.53 & 1.27 & 8,056 & 1.11 & 185.01 & 17.35 & 37.39 & 4 \\

5.53 & 1.27 & 7,382 & 1.11 & 185.01 & 17.35 & 40.80 & 4 \\

5.53 & 1.26 & 6,673 & 1.11 & 185.01 & 17.35 & 45.14 & 4 \\

5.53 & 1.26 & 6,356 & 1.11 & 185.01 & 17.35 & 47.40 & 4 \\

\bottomrule
\end{tabular}

\end{table}

    
    Keeping in view previous studies \cite{antonino2013scale}, we have considered the following parameters for data collection:
\begin{itemize}
    \item Stator Currents
    \item Rotor Currents
    \item Input Power
    \item Slip/Rotor Frequency
    \item Rotor Speed
    \item Efficiency
\end{itemize}

A sample of the generated dataset is shown in \textit{Table II}. We divided the data into training and test set with 70-30\% ratio. Which implies that 70\% of the collected data points were used to train the model. This is done to ensure that our model is well trained under all possible conditions that could occur in real life scenarios. Remaining 30\% of the data is tested upon to observe if any faults are detected in the model. 

We shall now see the detail of each fault induced in the motor.

\section{\textbf{Analysis of Motor Faults}}
Let us now analyze the various faults that we induced in the motor. We will see the characteristics of the motor that are being effected with particular faults.

    \subsection{Healthy Motor}
    To analyze the behavior of motor with faults, we first need to understand the characteristics of a healthy motor. 

\begin{figure}[htp]
    \centering
    \includegraphics[width=7cm]{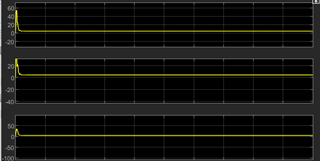}
       \caption{Stator Currents of Healthy Motor} 
\end{figure}
\vspace*{-1.5\baselineskip}
\begin{figure}[htp]
    \centering
    \includegraphics[width=7cm]{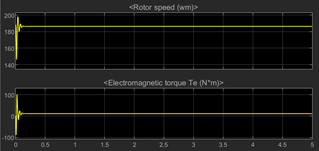} 
    \caption{Characteristics of Healthy Motor}
 
\end{figure}

    In \textit{figure 2 and 3}, it can be seen the normal current is flowing in the motor. In the start, current reaches a spike which is due to the inertia of the rotor which required excessive current to start the motor. As soon as the motor gains speed, we can observe uniform output.

    \subsection{Open Circuit Fault}
    When stator terminals were open circuited, the current flowing in the stator terminals dropped to zero, corresponding effecting the rotor speed and electromagnetic torque. This anomalous behavior was observed in open stator terminal as shown in \textit{figure 4 \& 5.}

\begin{figure}[htp]
    \centering
    \includegraphics[width=6.5cm]{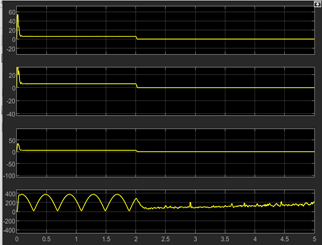}
       \caption{Stator Currents of Open-Circuited Motor} 
\end{figure}
\vspace*{-1.5\baselineskip}
\begin{figure}[htp]
    \centering
    \includegraphics[width=6.5cm]{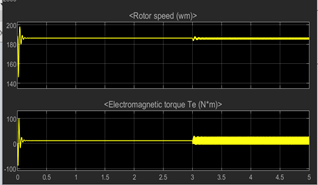} 
    \caption{Characteristics of Open-Circuited Motor}

\end{figure}
  \vspace*{-0.8\baselineskip}
    \subsection{Short Circuit Fault}
    When we short circuited the input terminals, high current started flowing in the rotor. Abnormal power was flowing in the motor. In the practical life motor can’t bear this much increase in current and leads to breakdown of motor.
\begin{figure}[htp]
    \centering
  \includegraphics[width=6.5cm]{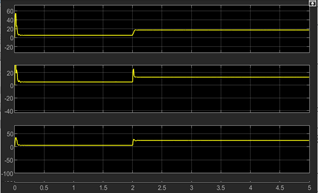}
       \caption{Stator Currents of Short-Circuited Motor  } 
\end{figure}
\vspace*{-1.5\baselineskip}
\begin{figure}[H]
    \centering
    \includegraphics[width=6.5cm]{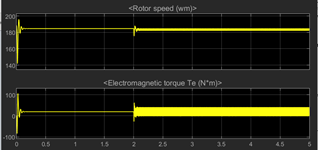} 
    
    \caption{Characteristics of Short-Circuited Motor}
 
\end{figure}

    \subsection{Motor Overload Fault}
    When the motor is load above its rated value,  input current increase due to increase in input power to compensate the energy requirements of overload. Rotor current frequency also increased, this is because of increase in slip which is directly proportional to rotor frequency. Due to overload, rotor speed was decreased and electromagnetic torque was increased as it is inversely proportional to rotor speed. 

\begin{figure}[htp]
    \centering
    \includegraphics[width=7cm]{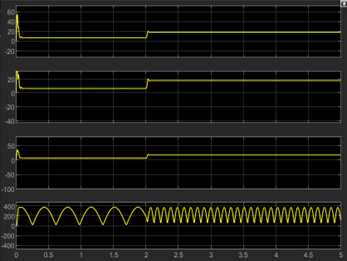}
       \caption{Stator Currents of Overloaded Motor} 
\end{figure}
\begin{figure}[htp]
    \centering
    \includegraphics[width=7cm]{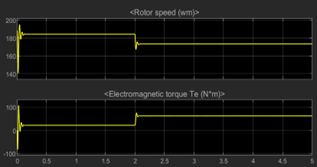}
    \caption{Characteristics of Overloaded Motor}
\end{figure}
\begin{figure}[htp]
    \centering
    \includegraphics[width=7cm]{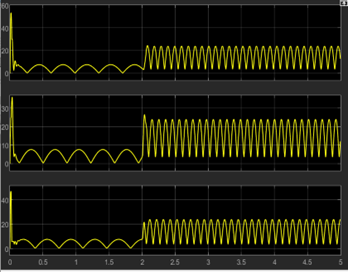} 
       \caption{Rotor Currents of Overloaded Motor} 
 
\end{figure}

    \subsection{Broken Rotor Bar Fault}
    When rotor bar is broken, harmonics are added in the signal as discussed previously. This results in distorted sinusoid  as shown in \textit{figure 11 and 12}. \cite{karvelis2015symbolic} \cite{rotorfault}

    \begin{figure}[htp]
    \centering
    \includegraphics[width=7cm]{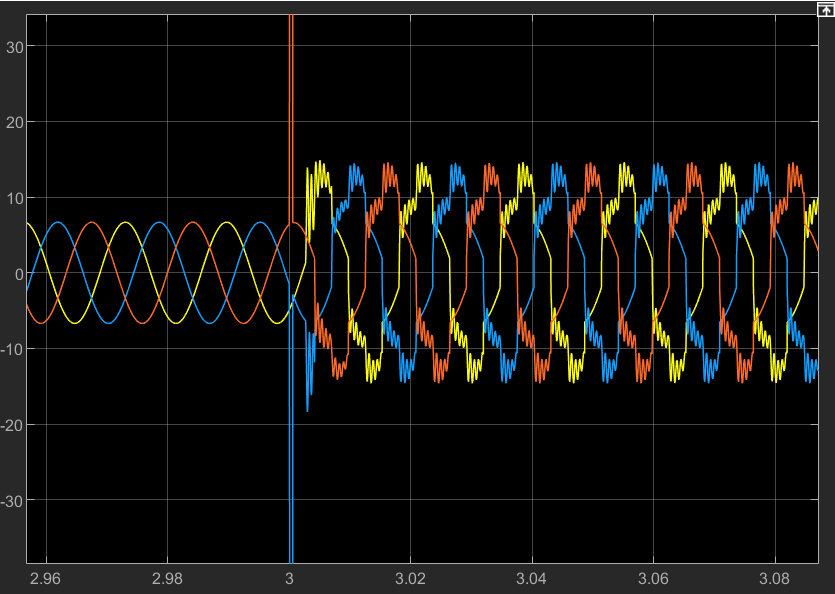}
       \caption{Stator Currents of Motor with Broken Rotor Bar} 
\end{figure}
\begin{figure}[H]
    \centering
    \includegraphics[width=7cm]{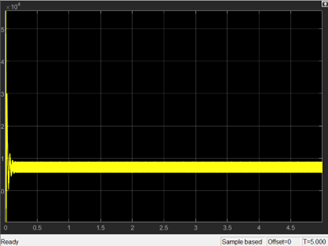} 
    \caption{Input Power of Motor with Broken Rotor Bar}
 
\end{figure}

    \subsection{Application of Signal Processing}
    Literature reveals various signal processing techniques for the extraction of features from our dataset such as wavelet transform, fourier transform etc. \cite{gao2015feature} \cite{keskes2015recursive} Here we have applied fourier transform to analyze the frequency domain of our motor features as shown in \textit{figure 13} .

\begin{figure}[htp]
    \centering
    \includegraphics[width=\linewidth]{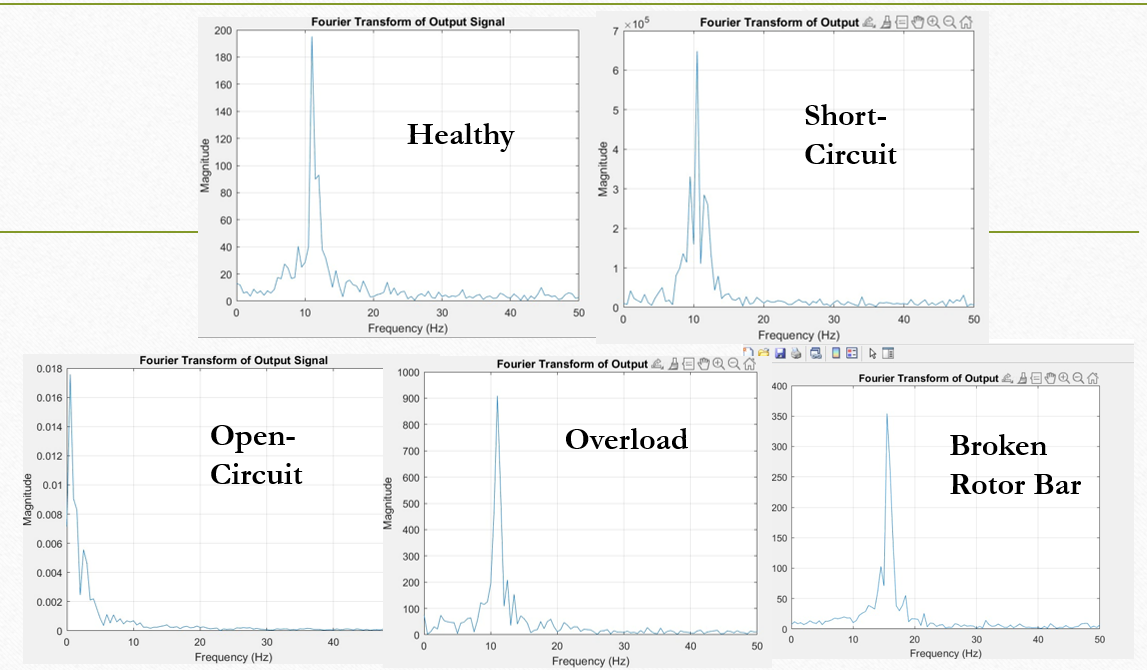}
    \caption{Fourier Transform of Stator Current}
    \label{fig:STM3}
\end{figure}

We can observe that for short-circuit the magnitude of the frequency components is extremely high. on the other hand, for open-circuit, there are components at very low frequency. For overload, their is also an increase in the magnitude due to excess drawing of current. Finally, we observe that for a broken rotor bar fault, there are additional frequency components added in the circuit causing distortion in pure sinusoid as discussed in previous section. Now we will see how these features were classified using machine learning. \cite{boukra2013statistical}\cite{wang2014multi}

\section{\textbf{Application of Machine Learning}}
The Classification Learner app in MATLAB was utilized for training machine learning models on the preprocessed dataset. \cite{matic2012support}\cite{chen2012prediction}

\textit{Figure 14} shows the plot matrix of the dataset. This give us a picture of the inter feature relationships. 
\begin{figure}[htp]
    \centering
    \includegraphics[width=\linewidth]{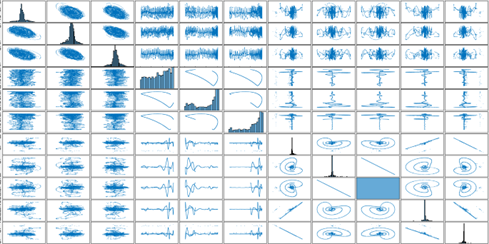}
    \caption{Plot Matrix}
    \label{figure:PM}
\end{figure}

The plot matrix gives machine learning engineers an idea about the nature of dataset and helps them to decide whcih model will perform better on a particular dataset. Moreover it also gives the information about the redundant features so that they may be removed using Principle Component analysis (PCA) \cite{plotmatrix}. Seeing the plot matrix in our case, we can observe that no feature is redundant so we have disabled PCA before the training process. 

\setcounter{subsection}{0}

    \subsection{Performance comparision of ML Models}
    Various algorithms, including Decision Trees, Support Vector Machines, Logistic Regression and Naive Bayes, were evaluated to determine the most effective model for fault classification. \cite{lei2016intelligent}\cite{murphey2006model} The models were assessed using metrics such as accuracy, precision, recall, and F1 score to gauge their performance. The results are shown in the \textit{figure 15}: 
\begin{figure}[htp]
    \centering
    \includegraphics[width=\linewidth]{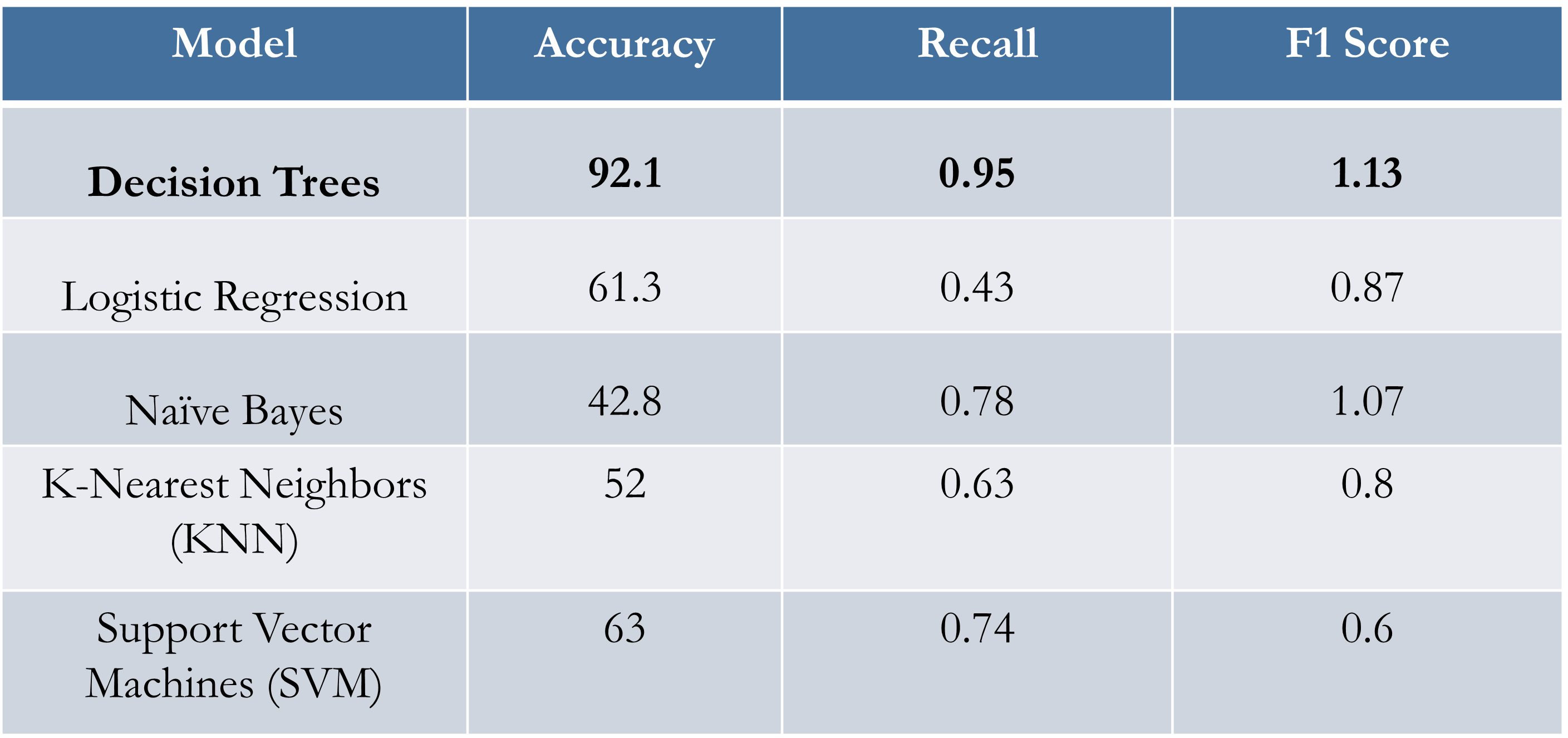}
    \caption{Performance of Machine Learning Models}
    \label{fig:Atmeg6}
\end{figure}

    The Decision Tree algorithm emerged as the most promising candidate, demonstrating superior accuracy and reliability in distinguishing between fault classes. \textit{Figure 16} Shows the confusion matrix of the desicion tree algorithm. It shows the accuracy of the model on individual faults. Some of the faults are very distinctive so their detection accuray is 100\%.
\begin{figure}[htp]
    \centering
    \includegraphics[width=7.8cm]{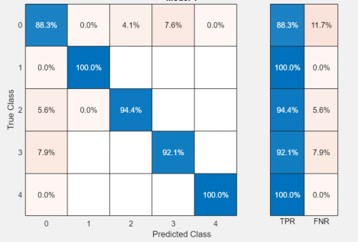}
    \caption{Confusion Matrix of Decision Tree Algorithm}
    \label{fig:Atmega6}
\end{figure}
    
    We shall now see the reason behind such high accuracy of decision tree algorithm as opposed to other popular models.

    \subsection{Decision Tree Algorithm}
    The Decision Tree algorithm is a versatile and interpretable machine learning technique widely used for classification tasks.
    
    It operates by recursively partitioning the input space based on the most significant features, creating a tree-like structure of decision nodes as shown in \textit{figure 17}.

\begin{figure}[htp]
    \centering
    \includegraphics[width=7.8cm]{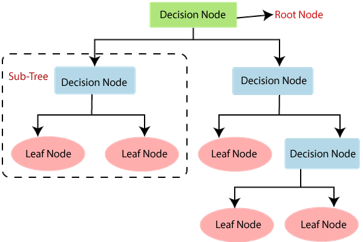}
    \caption{Decision Tree Algorithm}
    \label{fig:Atmega1}
\end{figure}

    \subsection{Enhanced performance of Decision Tree algorithm}
    \textbf{Feature Importance:} Decision Trees inherently rank features based on their importance in classifying faults. This aids in identifying the critical factors contributing to fault detection.
    
    \textbf{Adaptability to Complex Data:} The ability of Decision Trees to adapt to complex and non-linear relationships in the dataset aligns well with the intricate nature of induction motor fault patterns.
    
    \textbf{Robustness:} Decision Trees are robust to outliers and noise in the dataset, which is crucial when dealing with real-world data that may contain uncertainties.

    Hence, the Decision Tree algorithm, post-training on the dataset, exhibited commendable accuracy and efficiency in diagnosing induction motor faults. 

        
        





\section{\textbf{Conclusion and Future Work}}
 This paper aimed to address the complexities of detecting and classifying faults in induction motors, with a focus on practical applicability. Key achievements include the development of a comprehensive Simulink model, capturing nuanced motor behavior under various fault conditions. Rigorous signal processing techniques enhanced the dataset's discriminatory power, and machine learning integration introduced a data-driven approach to fault diagnosis. The Decision Tree algorithm stood out as the preferred choice, demonstrating interpretability, adaptability to non-linear relationships, and robustness against noise, making it highly effective for fault classification in this context.

The future works involve exploring advanced signal processing methods for extracting relevant features from motor operation data, such as time-domain and frequency-domain analysis and developing techniques for continuous condition monitoring of induction motors as well as quantifying the severity of detected faults, providing insights into the urgency and extent of required maintenance.

\section{\textbf{Acknowledgments}}

We express my sincere appreciation to those who played pivotal roles in the success of this paper. Special thanks to our lab instructor, Mr. Yasir Rizwan, for their invaluable guidance. I am also grateful to faculty advisor, Dr. Farid Gul for their continuous support. Collaborators and peers significantly enriched the research with their contributions and enthusiasm. The resources provided by NUST were instrumental. My gratitude extends to friends and family for their unwavering encouragement.

\end{document}